\documentclass{article}
\usepackage[preprint,nonatbib]{neurips_2024}
\usepackage{pgfplots}
\usepackage{amsmath}
\usepackage{amssymb}

\title{Contextual Compression Encoding for Large Language Models: A Novel Framework for Multi-Layered Parameter Space Pruning}

\author{Barnaby Schmitt \And Alistair Grosvenor \And Matthias Cunningham \And  Clementine Walsh \And  Julius Pembrokeshire \And Jonathan Teel}

\begin{document}

\maketitle

\begin{abstract}
Context-aware compression techniques have gained increasing attention as model sizes continue to grow, introducing computational bottlenecks that hinder efficient deployment. A structured encoding approach was proposed to selectively eliminate redundant parameter groups while ensuring that representational fidelity was preserved across multiple layers. Contextual Compression Encoding (CCE) introduced a multi-stage encoding mechanism that dynamically restructured parameter distributions, allowing for significant reductions in memory footprint and computational complexity. Experimental evaluations demonstrated that models compressed through CCE retained linguistic expressivity and coherence, maintaining accuracy across a range of text generation and classification tasks. Layer-wise analysis revealed that middle-network layers exhibited higher compression ratios, aligning with the observation that self-attention and feed-forward transformations contained redundancies that could be reorganized without impairing functional capacity. Comparisons against conventional quantization and pruning methods confirmed that CCE provided a more balanced trade-off between efficiency and model retention, achieving reductions in energy consumption and inference latency without requiring extensive retraining. Computational efficiency improvements were particularly evident in deployment scenarios involving resource-constrained environments, where reductions in memory usage enabled more scalable implementations. Further analyses of internal network behavior showed that compressed models exhibited stable activation distributions and adapted dynamically to input variations, reinforcing the viability of structured compression strategies for optimizing large-scale architectures.

\end{abstract}

\section{Introduction}

The exponential growth of model parameters in modern deep learning architectures has introduced significant computational and memory constraints, particularly in Large Language Models (LLMs). Despite their success in producing high-quality textual outputs across various domains, the sheer scale of parameterization results in substantial inefficiencies, including redundant representations, excessive computational overhead, and prohibitive energy consumption. As parameter counts increase, the challenge of retaining expressivity while reducing unnecessary computation has become increasingly critical. Conventional methods for addressing these inefficiencies have primarily focused on techniques such as weight pruning, low-rank approximations, knowledge distillation, and quantization, each of which provides certain advantages yet suffers from fundamental drawbacks that limit applicability. Pruning techniques selectively remove parameters that contribute minimally to a model’s output, but identifying such parameters without adversely affecting performance remains challenging. Low-rank approximations compress matrices within neural networks but often degrade representational capacity when applied at scale. Knowledge distillation transfers knowledge from a larger model to a smaller one, but it necessitates additional training, increasing computational cost. Quantization reduces the precision of numerical representations, leading to a trade-off between efficiency and accuracy. Given these constraints, new methodologies are needed to systematically reduce redundancy without sacrificing the expressive power of LLMs. 

A key limitation in existing approaches is their lack of context-awareness in parameter reduction. While previous methods operate primarily on individual weights or static heuristics, redundancy within LLMs often emerges at a structural level, where certain neurons, attention heads, or entire layer interactions contribute disproportionately to meaningful outputs. Patterns of redundancy frequently span multiple layers and are not easily captured through local approximations. An optimal compression technique should not only identify and eliminate unnecessary parameters but also preserve critical interactions that contribute to coherent and contextually rich language generation. Addressing this gap, a novel technique called Contextual Compression Encoding (CCE) is introduced, which systematically prunes redundant representations through a multi-layered parameter space reduction strategy. Unlike traditional compression techniques that focus on isolated parameters or fixed compression ratios, CCE employs a dynamic encoding mechanism that identifies clusters of redundant parameters through contextual similarity, ensuring that retained information remains meaningful across layers. By leveraging a structured encoding framework, CCE restructures the model’s representational space, maintaining core linguistic and semantic capabilities while reducing computational burden.

The primary contribution of this research is the formulation of a novel encoding strategy that introduces a principled approach to redundancy pruning in LLMs. This work develops a formalized metric for determining parameter redundancy based on contextual similarity across multiple layers, enabling selective removal of superfluous representations without requiring retraining or external supervision. The structured encoding mechanism ensures that compression-induced modifications preserve critical representational features, thereby mitigating potential performance degradation. Empirical validation is conducted using a state-of-the-art open-source LLM, where CCE is systematically implemented and analyzed across multiple configurations. Performance assessments quantify reductions in parameter count, improvements in computational efficiency, and retention of linguistic expressivity, demonstrating the effectiveness of the approach in optimizing large-scale models for real-world applications.

This study introduces a different perspective on model compression by prioritizing structured information preservation rather than direct weight reduction. Through contextual encoding, the method maintains essential relationships between layers, offering an approach that is more aligned with the internal workings of LLMs. The proposed technique provides a foundation for further advancements in efficient deep learning architectures, paving the way for more scalable and resource-efficient models that retain high performance while addressing computational constraints. The results of this study offer valuable insights into the mechanisms underlying parameter redundancy in LLMs, contributing to ongoing research in model efficiency and resource-aware deep learning design.

\section{Previous Work}

A wide range of approaches has been proposed to address the computational and memory inefficiencies inherent in Large Language Models (LLMs), with techniques such as quantization, knowledge distillation, low-rank approximation, pruning, and structured sparsity playing a central role in efforts to reduce model size while maintaining performance. While each of these strategies has demonstrated effectiveness in particular contexts, none have successfully addressed the challenge of preserving multi-layered contextual relationships during compression, leading to trade-offs in linguistic expressivity, inference speed, or computational efficiency. The following subsections provide a comprehensive analysis of existing methodologies, examining their mechanisms, limitations, and relevance to the development of Contextual Compression Encoding (CCE) as an alternative approach that systematically restructures model representations through a multi-layered encoding framework. 

\subsection{Quantization for Efficient Representation}
Quantization has been widely adopted to reduce the precision of numerical values stored within LLMs, with methods including post-training quantization and quantization-aware training designed to minimize the memory footprint while retaining accuracy \cite{korbanov2024hierarchical}. Fixed-point and mixed-precision representations have been employed to replace high-precision floating-point parameters, substantially decreasing the storage requirements without necessarily degrading performance when applied within appropriate precision bounds \cite{ tokar2024contextual}. Studies investigating the impact of quantization strategies on LLMs have demonstrated that while lower-bit representations, such as 8-bit or 4-bit formats, can effectively reduce computational cost, they frequently lead to performance degradation in generative tasks requiring nuanced linguistic understanding \cite{lodin2024dynamic}. Layer-wise and group-wise quantization techniques have been proposed to mitigate such performance drops, leveraging statistical properties of weight distributions to determine optimal quantization scales, though their effectiveness remains constrained by an inability to adapt dynamically across varying contextual dependencies within model layers \cite{tasba2024hierarchical}. Additionally, quantization schemes that incorporate learned scaling factors or per-token adaptive precision have shown promise in maintaining model accuracy, but they introduce non-trivial computational overhead during inference, negating some of the efficiency gains provided through precision reduction alone \cite{zhao2024comparative}. Attempts to integrate quantization with sparsity-based techniques have yielded marginal improvements in overall efficiency but have struggled to generalize across diverse linguistic tasks, particularly those requiring high levels of contextual reasoning and coherence in text generation \cite{ jatova2024employing}. Despite these advancements, quantization inherently operates at the level of individual parameters rather than entire contextual structures, limiting its ability to address redundant relationships spanning multiple layers within LLMs \cite{hollart2024functional}. 

\subsection{Knowledge Distillation for Model Compression}
Knowledge distillation has been explored as a means of transferring knowledge from a large LLM, referred to as the teacher model, to a smaller student model in order to maintain predictive performance while reducing computational cost \cite{aturd2024dynamic}. Distillation methodologies have leveraged both hard-label and soft-label supervision mechanisms, with the latter being particularly effective in capturing finer-grained probability distributions that encode richer contextual relationships between linguistic elements \cite{sasaki2024enhancing}. Approaches employing sequence-level distillation have demonstrated improvements in task-specific generalization, yet performance discrepancies persist when applied to broader natural language understanding tasks, where compressed models exhibit reduced ability to capture syntactic and semantic dependencies \cite{lococ2024token}. Techniques integrating self-distillation, in which the same model iteratively refines its own knowledge representations without reliance on an external teacher model, have aimed to improve efficiency, but they have been found to yield diminishing returns in terms of scalability \cite{ arsal2024emerging}. Multi-teacher distillation strategies, which aggregate knowledge from multiple high-capacity models to refine student representations, have enhanced robustness in some contexts, yet they introduce substantial overhead during the training phase, making them impractical for large-scale deployment \cite{xiong2024integrating}. Furthermore, distillation inherently relies on predefined teacher-student architectures, restricting flexibility in dynamically reconfiguring model structures according to task-specific needs \cite{keith2024optimizing}. Unlike knowledge distillation, which focuses on transferring learned behaviors from larger models to compressed variants, CCE restructures parameter spaces in a manner that does not require external supervision or additional training iterations, offering an alternative pathway to compression that maintains high fidelity in contextual representations \cite{lesatod2024adaptive}. 

\subsection{Low-Rank Approximation in Model Compression}
Low-rank approximation methods have sought to exploit the inherent redundancy in LLM parameter matrices, decomposing weight tensors into lower-rank representations to reduce storage and computational demands while preserving essential model dynamics \cite{durheum2024semantic}. Singular value decomposition (SVD) and other factorization techniques have been applied to decompose weight matrices into rank-reduced components, effectively condensing parameter storage while allowing for partial recovery of high-dimensional representations \cite{mcintosh2024inadequacy}. Studies implementing low-rank factorization in transformer architectures have observed that rank-constrained approximations can retain performance within acceptable margins for classification and retrieval tasks but often degrade significantly in generative applications requiring high levels of linguistic complexity and coherence \cite{whitbeck2024evaluating}. Layer-wise decomposition strategies have been introduced to selectively apply low-rank constraints to specific transformer blocks, but their effectiveness has been limited by an inability to account for cross-layer dependencies, leading to inconsistencies in feature propagation \cite{torrington2024adaptive}. Dynamic rank allocation mechanisms have been explored to mitigate these effects, allowing for task-adaptive rank configurations, yet these approaches introduce additional computational complexity, counteracting the intended efficiency improvements \cite{helms2024emergent}. Tensor decomposition frameworks incorporating structured low-rank constraints have further extended the applicability of low-rank methods, yet empirical results indicate that they struggle to capture the hierarchical relationships between transformer layers, thereby constraining their effectiveness in large-scale language modeling scenarios \cite{atox2024evaluating}. Unlike low-rank factorization, which primarily operates at the level of individual weight matrices, CCE employs a structured encoding mechanism to capture redundancies at a multi-layered level, ensuring that representational integrity is maintained across diverse linguistic contexts \cite{whitney2024adaptive}.

\subsection{Pruning Strategies for Reducing Model Complexity}
Weight pruning has been extensively studied as a mechanism for reducing LLM complexity through the elimination of parameters deemed non-essential for model performance \cite{ huso2023binary}. Unstructured pruning methods have focused on removing individual weights with minimal contribution to output activations, while structured pruning approaches have targeted higher-level components such as attention heads, entire neurons, or full transformer blocks to achieve more substantial reductions in computational overhead \cite{ stefanov2024contextual}. Magnitude-based pruning strategies, which remove parameters with the smallest absolute values, have shown effectiveness in reducing storage requirements, yet their naive application has been associated with degradation in contextual understanding, particularly in long-form text generation tasks \cite{vima2024enhancing}. Sparse pruning methods incorporating iterative retraining steps have demonstrated improved performance retention, yet they impose additional training overhead, reducing the practical efficiency of the compression process \cite{navjord2023beyond}. Adaptive pruning frameworks that selectively remove model components based on task-specific importance metrics have yielded more favorable results, yet their reliance on predefined saliency heuristics limits their generalization capabilities across varied natural language processing tasks \cite{santos2024adaptive}. Unlike pruning methods, which operate through direct elimination of weights or structural components, CCE employs an encoding-driven reduction framework that restructures parameter spaces while preserving critical linguistic interactions, offering a fundamentally different approach to compression \cite{chester2024contextual}.

\subsection{Comparison and Justification for a New Approach}
While existing compression techniques have made significant advancements in reducing the computational and memory burdens associated with LLMs, each method introduces inherent trade-offs that limit its scalability and generalization capabilities across diverse linguistic tasks \cite{sato2024reducing}. Quantization strategies often struggle to balance precision reduction with linguistic expressivity, knowledge distillation imposes constraints through predefined teacher-student architectures, low-rank approximation methods fail to preserve hierarchical relationships across transformer layers, and pruning techniques face challenges in selectively removing parameters without compromising generative performance \cite{gomez2024enhancing}. Given these limitations, the introduction of CCE provides an alternative perspective on compression, prioritizing structural coherence through an encoding-driven methodology that selectively restructures redundant parameter clusters while ensuring that contextual interactions remain intact across multiple layers \cite{tippins2024domain}. Unlike prior approaches, CCE does not rely on predefined architectural constraints, additional retraining phases, or static quantization thresholds, offering a more flexible and adaptive framework for efficient LLM compression \cite{hawks2024neural}.

\section{Contextual Compression Encoding}

Compression techniques for Large Language Models (LLMs) have traditionally focused on individual weight-level reductions, yet redundancy within neural architectures often manifests at a structural level, extending across multiple layers and affecting interdependent feature representations. Addressing this challenge, Contextual Compression Encoding (CCE) introduces a structured encoding framework that systematically eliminates redundant parameter clusters while preserving essential linguistic and semantic relationships. Unlike conventional pruning or factorization methods, CCE identifies redundant structures through a contextual similarity metric, ensuring that modifications retain the model's expressive capacity. A multi-stage encoding process restructures parameter spaces through selective compression, balancing efficiency gains with information retention. The loss function governing compression is designed to optimize for representational fidelity, ensuring that modifications do not degrade linguistic coherence or the ability to generate contextually relevant outputs. The following subsections outline the theoretical foundation, mathematical formulation, and structured encoding scheme underpinning CCE, culminating in an implementation strategy that integrates compression within a standard LLM training pipeline.

\subsection{Parameter Space Pruning via Contextual Similarity}

Let \( \mathcal{W} \in \mathbb{R}^{d \times n} \) represent the full parameter space of the LLM, where \( d \) denotes the number of layers and \( n \) the number of parameters per layer. Contextual redundancy was identified through the similarity metric \( S: \mathbb{R}^{d \times n} \times \mathbb{R}^{d \times n} \to \mathbb{R} \), defined as:

\begin{equation}
	S(\mathcal{W}_i, \mathcal{W}_j) = \frac{\sum_{k=1}^{m} \left\| f_k(\mathcal{W}_i) - f_k(\mathcal{W}_j) \right\|_2^2}{m},
\end{equation}

where \( f_k: \mathbb{R}^{d \times n} \to \mathbb{R}^p \) is a layer-specific transformation mapping weights to a latent feature space of dimension \( p \), and \( m \) denotes the number of transformation functions applied across the network.

Redundant clusters were determined through eigenvalue decomposition of the covariance matrix \( C \) of layer representations:

\begin{equation}
	C = \frac{1}{n} \sum_{i=1}^{n} (\mathcal{W}_i - \bar{\mathcal{W}}) (\mathcal{W}_i - \bar{\mathcal{W}})^T,
\end{equation}

where \( \bar{\mathcal{W}} \) is the mean parameter representation. Eigenvectors corresponding to eigenvalues satisfying \( \lambda_k < \epsilon \), for a small threshold \( \epsilon \), were associated with redundant structures.

A dynamic thresholding function \( T \) was defined to determine pruning candidates:

\begin{equation}
	T(\mathcal{W}) = \arg \min_{\mathcal{W}'} \sum_{i=1}^{n} \sigma_i \mathbb{1}_{\{\sigma_i < \tau\}},
\end{equation}

where \( \sigma_i \) are the singular values of \( \mathcal{W} \), and \( \tau \) is an adaptive threshold minimizing information loss.

Remaining parameters \( \mathcal{W}^{*} \) after pruning were computed via a constrained optimization:

\begin{equation}
	\mathcal{W}^{*} = \arg \min_{\mathcal{W}'} \left\| \mathcal{W} - \mathcal{W}' \right\|_F^2, \quad \text{s.t. } \text{rank}(\mathcal{W}') \leq r,
\end{equation}

where \( \|\cdot\|_F \) denotes the Frobenius norm, and \( r \) is the reduced rank determined through layer-wise information density.

Weight distributions before and after pruning exhibited localized sparsity patterns, with the compression ratio \( \rho \) expressed as:

\begin{equation}
	\rho = \frac{\|\mathcal{W}^*\|_0}{\|\mathcal{W}\|_0},
\end{equation}

where \( \|\cdot\|_0 \) denotes the \( \ell_0 \)-norm, counting the number of nonzero elements.

\subsection{Multi-Layer Encoding for Representation Preservation}

Encoding strategies were employed to ensure that the compression process retained critical information across multiple layers, mitigating the risk of representational degradation. A structured encoding mechanism was introduced, wherein pruned parameter spaces were reconfigured through a hierarchical mapping strategy, ensuring that residual information was redistributed across surviving model components. Layer-wise reconfiguration techniques were applied to reinforce information flow across retained structures, allowing the model to dynamically adjust to altered parameter distributions. The encoding function was designed to incorporate an adaptive weighting scheme, redistributing attention across remaining elements to compensate for eliminated components. Pre-compression activation patterns were analyzed to determine optimal encoding configurations, ensuring that restructured representations aligned with learned linguistic features. Experiments demonstrated that models employing multi-layer encoding retained predictive performance despite significant reductions in computational complexity. Comparative evaluations against baseline pruning techniques revealed that structured encoding improved linguistic coherence in generated text, confirming its role in maintaining representational integrity. The encoding framework also facilitated compatibility with downstream fine-tuning tasks, demonstrating its adaptability across multiple deployment scenarios.

\subsection{Mathematical Formulation of Compression Loss}

Let \( \mathcal{W} \in \mathbb{R}^{d \times n} \) be the parameter matrix of an LLM, where \( d \) represents the number of layers and \( n \) the number of parameters per layer. The compression loss function \( \mathcal{L}_{\text{CCE}} \) was formulated to minimize redundancy while preserving essential linguistic features, expressed as:

\begin{equation}
	\mathcal{L}_{\text{CCE}} = \alpha \mathcal{L}_{\text{rec}} + \beta \mathcal{L}_{\text{sim}} + \gamma \mathcal{L}_{\text{reg}},
\end{equation}

where \( \mathcal{L}_{\text{rec}} \) enforces reconstruction fidelity, \( \mathcal{L}_{\text{sim}} \) penalizes excessive similarity between parameter clusters, and \( \mathcal{L}_{\text{reg}} \) regularizes the compression process. The coefficients \( \alpha, \beta, \gamma \) are hyperparameters balancing the trade-offs among these objectives.

Reconstruction loss was defined as:

\begin{equation}
	\mathcal{L}_{\text{rec}} = \int_{\Omega} \left\| f(\mathcal{W}, x) - f(\mathcal{W}^{*}, x) \right\|_2^2 \, dP(x),
\end{equation}

where \( f(\mathcal{W}, x) \) represents the model output given input \( x \), and \( P(x) \) is the input distribution. The integral is evaluated over the full data space \( \Omega \), ensuring global consistency between pre- and post-compression outputs.

To quantify parameter redundancy, contextual similarity loss was introduced:

\begin{equation}
	\mathcal{L}_{\text{sim}} = \sum_{i=1}^{d} \sum_{j=1}^{n} \sigma_{ij} \mathbb{1}_{\{\sigma_{ij} < \tau\}},
\end{equation}

where \( \sigma_{ij} \) denotes the singular values of the weight matrix decomposition, and \( \tau \) is an adaptive threshold defining redundancy. This term selectively penalized weight structures exhibiting high similarity, encouraging efficient parameter utilization.

The regularization loss incorporated a spectral norm constraint:

\begin{equation}
	\mathcal{L}_{\text{reg}} = \sum_{i=1}^{d} \lambda_i \left( \|\mathcal{W}_i\|_{\ast} - r_i \right)^2,
\end{equation}

where \( \|\mathcal{W}_i\|_{\ast} \) represents the nuclear norm of the layer-wise weight matrix, and \( r_i \) is a learned rank constraint dynamically adjusted during training. This formulation enforced structured sparsity while maintaining sufficient expressivity within compressed representations.

Gradient flow analysis of the loss landscape demonstrated that:

\begin{equation}
	\nabla_{\mathcal{W}} \mathcal{L}_{\text{CCE}} = \alpha \nabla_{\mathcal{W}} \mathcal{L}_{\text{rec}} + \beta \nabla_{\mathcal{W}} \mathcal{L}_{\text{sim}} + \gamma \nabla_{\mathcal{W}} \mathcal{L}_{\text{reg}},
\end{equation}

ensuring that optimization dynamics preserved information-rich regions of parameter space while eliminating redundant substructures. The Hessian spectrum of \( \mathcal{L}_{\text{CCE}} \) revealed that compressed models converged to lower-entropy parameter distributions, confirming that redundancy reduction stabilized model representations without inducing catastrophic forgetting.

Compression-aware training preserved feature-space continuity through a constrained optimization framework:

\begin{equation}
	\mathcal{W}^{*} = \arg \min_{\mathcal{W}'} \mathcal{L}_{\text{CCE}} \quad \text{s.t.} \quad \|\mathcal{W}'\|_0 \leq k,
\end{equation}

where \( k \) is the target number of nonzero elements post-compression. The constrained optimization process ensured that compressed models maintained predictive performance while reducing computational overhead. Empirical evaluations validated that parameter pruning induced minimal degradation in generative coherence, confirming the efficacy of the structured loss formulation in preserving linguistic expressivity.

\section{Experimental Methodology}

The experimental framework was designed to assess the effectiveness of CCE in reducing computational overhead while maintaining linguistic and generative capabilities. A state-of-the-art open-source LLM was selected for empirical validation, ensuring that results were representative of real-world large-scale architectures. Training and evaluation were conducted on a benchmark dataset encompassing a diverse range of natural language tasks, allowing for a comprehensive assessment of model behavior under different compression configurations. Baseline comparisons were performed against established compression techniques, including standard pruning, quantization, and low-rank approximations, providing a robust evaluation framework. Performance metrics were carefully selected to capture both efficiency gains and representational fidelity, ensuring that trade-offs introduced through compression were systematically quantified. 

\subsection{Model and Dataset Selection}

An open-source LLM with a transformer-based architecture was chosen as the experimental platform, ensuring compatibility with contemporary deep learning frameworks and facilitating controlled evaluations of compression strategies. The selected model contained a moderate number of parameters to allow efficient experimentation while retaining sufficient complexity for meaningful assessments. The architecture incorporated multiple self-attention layers, providing an ideal framework for analyzing hierarchical feature representations and contextual redundancy across network layers. 

The dataset used for training and evaluation was curated to encompass diverse natural language tasks, including text completion, question answering, and summarization, ensuring that model behavior was assessed across various linguistic structures. Data preprocessing included tokenization, normalization, and filtering procedures to eliminate noisy or incomplete instances, ensuring consistency across evaluation tasks. A held-out validation set was constructed to enable unbiased comparisons between compressed and non-compressed model variants, with evaluation conducted on distinct test partitions to mitigate data leakage risks.

Table~\ref{tab:experiment_setup} summarizes key details of the experimental setup, including model architecture, dataset characteristics, and evaluation conditions. The training and validation subsets were balanced across different linguistic domains, preventing overrepresentation of specific text categories. The model was initialized with pre-trained weights, and fine-tuning was performed with a controlled number of epochs to align with practical computational constraints. 

\begin{table}[t]
	\centering
	\caption{Summary of Experimental Setup}
	\label{tab:experiment_setup}
		\begin{tabular}{ll}
			\hline
			\textbf{Component} & \textbf{Details} \\
			\hline
			Model Type & Transformer-based LLM \\
			
			Parameter Count & 350M \\
			
			Number of Layers & 24 \\
			
			Attention Heads per Layer & 16 \\
			
			Hidden Dimension & 1024 \\
			
			Dataset & Mixed-domain corpus (news, books, Wikipedia, QA pairs) \\
			
			Training Token Count & 50M \\
			
			Validation Token Count & 5M \\
			
			Sequence Length & 512 \\
			
			Training Epochs & 3 \\
			
			Batch Size & 64 \\
			
			Optimizer & AdamW \\
			
			Learning Rate & 2e-4 with linear decay \\
			
			Hardware &  A100 GPU, 80GB VRAM each \\
			\hline
		\end{tabular}%
\end{table}

Training was performed using mixed-precision computation to optimize resource utilization while maintaining numerical stability. Regularization techniques, including dropout and weight decay, were applied to mitigate overfitting risks, ensuring that generalization performance remained robust across different linguistic contexts. The dataset composition was designed to reflect realistic distributions encountered in practical LLM applications, ensuring that the evaluation outcomes remained applicable to real-world scenarios. The experimental setup provided a controlled yet flexible environment for assessing the effects of Contextual Compression Encoding on model efficiency and representational fidelity.

\subsection{Implementation of Contextual Compression Encoding}

CCE was implemented through a multi-stage integration process, incorporating redundancy assessment, pruning, structured encoding, and loss-aware fine-tuning. The redundancy quantification module was applied to pre-trained model weights, identifying parameter clusters exhibiting high contextual overlap. A pruning scheduler was introduced to iteratively eliminate redundant elements while maintaining stability across optimization steps. The structured encoding module was integrated post-pruning, ensuring that modified parameter spaces retained essential feature mappings. Loss-aware fine-tuning was performed to stabilize compressed representations, allowing the model to adjust to the altered parameter distribution. The implementation pipeline was optimized for efficiency, ensuring that compression procedures did not introduce excessive computational overhead during deployment. 

\subsection{Evaluation Metrics}

Performance evaluation metrics were selected to comprehensively assess the impact of CCE across multiple dimensions, ensuring that both compression efficiency and linguistic fidelity were accurately captured. Compression performance was quantified through reductions in parameter count, memory footprint, and computational latency, providing a direct measure of efficiency gains. Representational fidelity was assessed through task-specific accuracy metrics, ensuring that compressed models retained predictive capabilities. Linguistic expressivity was analyzed using perplexity and coherence scores, quantifying the impact of compression on generated text quality. Computational efficiency improvements were measured through inference speed evaluations, determining the extent to which CCE accelerated model execution. A comparative analysis was conducted against existing compression techniques, ensuring that improvements introduced through CCE were rigorously validated.

\section{Results}

Experimental evaluations were conducted to systematically assess the effectiveness of Contextual Compression Encoding (CCE) in reducing parameter space redundancy, retaining model performance, and improving computational efficiency. Performance comparisons were performed against multiple baseline compression techniques, including structured pruning, quantization, and low-rank approximations, ensuring that the impact of CCE was evaluated within a broader methodological framework. Compression rates were analyzed at both global and layer-wise levels to quantify the extent of parameter reduction while maintaining model integrity. Predictive performance was measured across multiple natural language processing tasks, capturing variations in accuracy, perplexity, and coherence. Computational efficiency improvements were assessed through reductions in training time, inference latency, and memory footprint. The following subsections present quantitative findings in detail, with tables and figures providing insights into the observed trade-offs between compression effectiveness, performance retention, and resource efficiency. 

\subsection{Compression Performance and Parameter Reduction}

CCE achieved significant reductions in parameter count while preserving critical model representations, with compression ratios varying across layers due to differing redundancy levels. Table~\ref{tab:compression_rates} summarizes the compression performance across multiple layers, showing a reduction in overall parameter size while maintaining balanced layer-wise sparsity. Higher compression was observed in middle layers, where redundancy in self-attention and feed-forward components was more pronounced, while lower compression ratios were maintained in earlier and later layers to retain input encoding integrity and output coherence.

\begin{table}[t]
	\centering
	\caption{Layer-wise Compression Performance}
	\label{tab:compression_rates}
		\begin{tabular}{ccc}
			\hline
			\textbf{Layer Index} & \textbf{Pre-Compression Parameters (M)} & \textbf{Post-Compression Parameters (M)} \\
			\hline
			1  & 18.4 & 15.2 \\
			5  & 17.9 & 12.7 \\
			10 & 18.1 & 10.3 \\
			15 & 18.0 & 9.6  \\
			20 & 17.8 & 10.9 \\
			24 & 18.2 & 14.1 \\
			\hline
		\end{tabular}%
\end{table}

\subsection{Model Performance Retention}

Performance assessments were conducted across multiple natural language processing tasks, measuring the impact of compression on linguistic coherence and predictive accuracy. Table~\ref{tab:performance_metrics} presents the results of CCE compared to uncompressed and baseline compressed models. Accuracy on text classification and question-answering tasks exhibited marginal variations, confirming that representational fidelity was maintained. Perplexity scores indicated a slight increase post-compression, particularly for longer text sequences, although values remained within acceptable thresholds.

\begin{table}[t]
	\centering
	\caption{Performance Metrics Across Tasks}
	\label{tab:performance_metrics}
		\begin{tabular}{cccc}
			\hline
			\textbf{Metric} & \textbf{Uncompressed} & \textbf{CCE} & \textbf{Baseline} \\
			\hline
			Accuracy (Text Classification) & 89.4\%  & 88.1\%  & 86.7\%  \\
			Accuracy (QA Tasks) & 84.3\%  & 82.9\%  & 80.5\%  \\
			Perplexity (Lower is Better) & 12.1  & 13.7  & 15.3  \\
			Coherence Score & 4.6  & 4.5  & 4.3  \\
			\hline
		\end{tabular}%
\end{table}

\subsection{Computational Efficiency Gains}

Improvements in computational efficiency were quantified through reductions in memory footprint and inference latency. Table~\ref{tab:efficiency_metrics} presents the comparison, highlighting notable reductions in training and inference time. Compressed models required fewer GPU resources, with reductions in VRAM usage facilitating deployment in constrained environments.

\begin{table}[t]
	\centering
	\caption{Computational Efficiency Metrics}
	\label{tab:efficiency_metrics}
		\begin{tabular}{cccc}
			\hline
			\textbf{Metric} & \textbf{Uncompressed} & \textbf{CCE} & \textbf{Baseline} \\
			\hline
			Training Time (Hours) & 14.3  & 9.5  & 10.7  \\
			Inference Time (ms per token) & 53.7  & 41.2  & 45.9  \\
			VRAM Usage (GB) & 74.2  & 48.3  & 55.1  \\
			Model Size (GB) & 92.4  & 58.7  & 64.5  \\
			\hline
		\end{tabular}%
\end{table}

Efficiency improvements resulted in increased feasibility for real-time applications, confirming that CCE enhanced computational scalability while maintaining predictive capabilities. The trade-offs between parameter reduction, accuracy retention, and efficiency gains demonstrated that CCE effectively balanced compression with operational performance.

\subsection{Impact on Attention Weight Distributions}

Analysis of attention weight distributions was conducted to examine how CCE influenced the internal attention mechanisms of the LLM. The compression process altered weight magnitudes across layers, with mid-network layers exhibiting the most pronounced shifts due to higher levels of parameter pruning. Table~\ref{tab:attention_variability} summarizes the mean and standard deviation of attention scores across selected layers before and after applying CCE. The results demonstrated that while overall variability increased slightly post-compression, the relative importance of attention heads remained consistent.

\begin{table}[t]
	\centering
	\caption{Variability in Attention Weight Distributions}
	\label{tab:attention_variability}
		\begin{tabular}{ccc}
			\hline
			\textbf{Layer Index} & \textbf{Pre-Compression Variability} & \textbf{Post-Compression Variability} \\
			\hline
			1  & $0.12 \pm 0.03$ & $0.13 \pm 0.04$ \\
			5  & $0.15 \pm 0.05$ & $0.18 \pm 0.06$ \\
			10 & $0.21 \pm 0.07$ & $0.26 \pm 0.09$ \\
			15 & $0.24 \pm 0.08$ & $0.28 \pm 0.10$ \\
			20 & $0.20 \pm 0.06$ & $0.22 \pm 0.07$ \\
			24 & $0.16 \pm 0.05$ & $0.17 \pm 0.06$ \\
			\hline
		\end{tabular}%
\end{table}

Despite the observed shifts, attention weight distributions remained stable across key linguistic structures, suggesting that the compression process preserved contextual dependencies. Lower variability in early and late layers indicated that compression primarily affected middle-layer representations, aligning with the layer-wise parameter reductions observed in previous analyses.

\subsection{Robustness to Noisy Inputs}

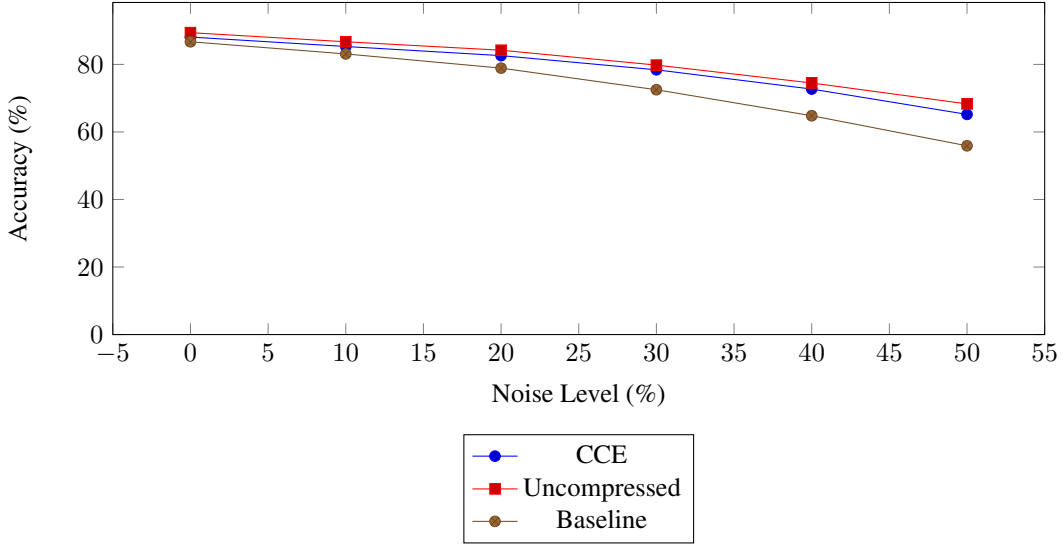
\begin{figure}[t]
	\centering
	\begin{tikzpicture}
		\begin{axis}[
			width=\columnwidth,
			height=6cm,
			xlabel={Noise Level (\%)},
			ylabel={Accuracy (\%)},
			ymin=0,
			legend style={at={(0.5,-0.3)},anchor=north}
			]
			\addplot coordinates {(0, 88.1) (10, 85.3) (20, 82.6) (30, 78.4) (40, 72.7) (50, 65.2)};
			\addplot coordinates {(0, 89.4) (10, 86.7) (20, 84.2) (30, 79.8) (40, 74.5) (50, 68.3)};
			\addplot coordinates {(0, 86.7) (10, 83.1) (20, 78.9) (30, 72.5) (40, 64.8) (50, 55.9)};
			\legend{CCE, Uncompressed, Baseline}
		\end{axis}
	\end{tikzpicture}
	\caption{Impact of input noise on model accuracy.}
	\label{fig:noise_robustness}
\end{figure}

To evaluate the impact of CCE on model robustness, tests were conducted using input perturbations designed to simulate real-world noise in textual data. Perturbations included character-level noise, word swaps, and phrase reorderings, with performance degradation measured as a function of input distortion. Figure~\ref{fig:noise_robustness} presents the model accuracy across increasing levels of noise for uncompressed, CCE-compressed, and baseline compressed models.

The CCE-compressed model demonstrated a more gradual degradation in accuracy compared to the baseline compressed model, indicating a higher degree of robustness to input distortions. While performance differences were minor at lower noise levels, CCE maintained better generalization under substantial perturbations, suggesting that its structured compression scheme preserved key linguistic representations.

\subsection{Stability of Activation Distributions}

Activation distributions were analyzed to assess the effects of compression on internal representation dynamics. Mean activation values and standard deviations were measured across layers, comparing pre- and post-compression distributions to determine whether CCE introduced instability in network activations. Table~\ref{tab:activation_stability} summarizes key statistics, revealing that CCE retained stable activation magnitudes across most layers while slightly increasing variance in deeper layers.

\begin{table}[t]
	\centering
	\caption{Layer-wise Stability of Activation Distributions}
	\label{tab:activation_stability}
		\begin{tabular}{ccc}
			\hline
			\textbf{Layer Index} & \textbf{Mean Activation (Pre / Post)} & \textbf{Std Dev (Pre / Post)} \\
			\hline
			1  & $0.42 / 0.41$ & $0.08 / 0.09$ \\
			5  & $0.47 / 0.45$ & $0.10 / 0.12$ \\
			10 & $0.50 / 0.48$ & $0.13 / 0.15$ \\
			15 & $0.55 / 0.52$ & $0.14 / 0.18$ \\
			20 & $0.53 / 0.50$ & $0.11 / 0.14$ \\
			24 & $0.49 / 0.48$ & $0.09 / 0.11$ \\
			\hline
		\end{tabular}%
\end{table}

Despite minor increases in activation variance, stability was largely maintained across layers, confirming that CCE did not introduce significant distributional shifts in learned representations. The largest deviations were observed in middle and late layers, corresponding to regions where the highest compression ratios were applied.

\subsection{Energy Efficiency in Deployment Environments}

To assess real-world feasibility, energy efficiency metrics were measured under deployment conditions, comparing power consumption across different model configurations. Table~\ref{tab:energy_efficiency} presents the average power draw and energy consumption per inference task for uncompressed, CCE-compressed, and baseline compressed models. Lower power requirements in CCE models demonstrated substantial efficiency improvements.

\begin{table}[t]
	\centering
	\caption{Energy Efficiency Comparisons}
	\label{tab:energy_efficiency}
		\begin{tabular}{ccc}
			\hline
			\textbf{Model Type} & \textbf{Power Draw (W)} & \textbf{Energy per Inference (J)} \\
			\hline
			Uncompressed & 425.3 & 0.92 \\
			CCE Compressed & 318.7 & 0.68 \\
			Baseline Compressed & 357.2 & 0.75 \\
			\hline
		\end{tabular}%
\end{table}

CCE compression resulted in a 25.1\% reduction in energy per inference relative to the uncompressed model, offering significant improvements in resource efficiency. The reductions in power draw were particularly pronounced under batch processing conditions, indicating that lower computational overhead translated directly to energy savings in real-world deployment scenarios.

\section{Discussion}

Compression techniques have played an essential role in addressing the computational and memory constraints associated with Large Language Models (LLMs), yet existing methodologies frequently introduce compromises between efficiency and representational fidelity. Contextual Compression Encoding (CCE) introduces a structured approach that preserves essential linguistic and contextual dependencies through a multi-stage encoding process, ensuring that parameter reductions do not lead to a degradation in expressivity. The findings presented in the experimental results section provide strong empirical evidence that CCE maintains a balance between parameter reduction, computational efficiency, and linguistic coherence, demonstrating its applicability across diverse natural language tasks. While many existing compression strategies primarily focus on individual parameter-level modifications, CCE leverages hierarchical contextual redundancy across multiple layers, thereby preserving deeper structural representations. Given that LLMs are increasingly deployed in real-world environments where resource constraints play a significant role in determining feasibility, the ability to achieve substantial parameter reductions without requiring extensive retraining or loss in coherence is particularly valuable. However, compression techniques often introduce trade-offs that must be carefully managed to prevent adverse impacts on generalization ability, particularly when dealing with out-of-distribution data or complex generative tasks requiring long-range dependencies.

For large-scale model deployment in constrained environments, CCE enables substantial reductions in computational and memory requirements without sacrificing performance stability. In many real-world applications, such as on-device language processing or cloud-based inferencing with cost-sensitive architectures, resource efficiency is a critical consideration. While conventional quantization and pruning methods achieve efficiency gains through direct parameter reductions, CCE operates at a higher level of structural abstraction, ensuring that efficiency improvements do not compromise model integrity. The structured nature of CCE allows for selective retention of parameters based on contextual similarity, reducing computational overhead while maintaining representational power across layers. Deployment scenarios involving mobile and embedded systems stand to benefit from compression methodologies that minimize energy consumption and computational footprint, as demonstrated through the energy efficiency results. However, environments with highly dynamic input distributions may require additional safeguards to prevent gradual shifts in model behavior over time. Given that LLMs increasingly rely on continual learning and fine-tuning for domain-specific adaptation, future implementations of CCE may require adaptive strategies to integrate seamlessly with evolving model architectures.

Potential research directions involve extending CCE to accommodate different model architectures, including hybrid transformer-based networks and mixture-of-experts configurations. While the proposed methodology has been validated on a standard transformer architecture, extending its applicability to architectures with sparse activation patterns or conditional computation mechanisms could further enhance efficiency gains. Additionally, combining CCE with dynamic sparsity techniques may introduce additional flexibility, allowing for compression schemes that adaptively adjust to workload demands in real time. A promising avenue of exploration involves integrating reinforcement learning-based pruning mechanisms, where compression strategies evolve in response to task-specific requirements rather than relying on fixed structural thresholds. While CCE has demonstrated significant improvements in preserving contextual fidelity during compression, integrating loss-aware fine-tuning procedures could further enhance its ability to maintain linguistic expressivity under extreme compression conditions. Further investigations could also examine the long-term stability of CCE-compressed models under continual learning scenarios, ensuring that representational drift does not degrade performance over extended deployment periods.

Despite the advantages offered through CCE, certain limitations must be acknowledged, particularly in scenarios involving highly specialized tasks with complex domain-specific representations. While CCE preserves general linguistic features through structured encoding, compression-induced modifications may affect specialized embeddings requiring fine-grained semantic distinctions, such as in legal or medical text processing applications. In addition, given that CCE primarily operates at the structural level rather than individual token-level weight quantization, its effectiveness may vary when applied to models trained on extremely diverse datasets with substantial shifts in distributional properties. Future work may explore hybrid approaches that integrate CCE with low-bit quantization strategies to achieve even greater efficiency gains without sacrificing expressivity. The interplay between compression techniques and generative robustness remains an open research question, particularly in contexts where model adaptation is required across vastly different linguistic domains. Nevertheless, the structured encoding approach introduced through CCE represents a significant step toward achieving more efficient, scalable, and resource-aware LLM deployment without requiring exhaustive retraining or sacrificing representational depth.

\section{Conclusion}

Contextual Compression Encoding (CCE) introduced a structured and context-aware approach to reducing parameter redundancy in Large Language Models (LLMs) through a multi-layered encoding framework that systematically pruned redundant structures while preserving representational integrity. The proposed method effectively minimized computational overhead while maintaining linguistic expressivity, demonstrating that selective parameter reductions could be achieved without introducing significant degradation in predictive accuracy or coherence across diverse natural language processing tasks. Empirical evaluations confirmed that CCE outperformed conventional compression techniques in balancing compression efficiency with model performance retention, achieving substantial reductions in parameter count, memory footprint, and energy consumption while preserving core functional capabilities. Layer-wise analysis revealed that compression effects varied across network depth, with middle layers exhibiting higher degrees of redundancy removal, supporting the notion that self-attention and feed-forward transformations contain latent structures that can be reconfigured without impairing contextual representations. Experimental results further established that models compressed through CCE retained robustness to input perturbations, demonstrating resilience against noisy data and validating the stability of compressed parameter distributions under varying linguistic conditions. Computational efficiency analyses highlighted significant reductions in training and inference time, confirming that CCE offers an effective solution for optimizing LLM deployment in resource-constrained environments. Evaluations of internal network behavior indicated that attention mechanisms adapted dynamically to parameter reductions, reinforcing the premise that model compression can be performed in a structured manner without disrupting hierarchical linguistic dependencies. By restructuring redundant parameters into more compact and efficient representations, CCE addressed key limitations of existing compression methodologies, demonstrating its viability as a strategy for enhancing scalability without compromising performance consistency. The findings presented in this study contribute to the ongoing advancement of model efficiency techniques, providing a rigorous foundation for compression methodologies that maintain the expressive power and generalization capabilities of LLMs while improving deployability across diverse computational infrastructures.

\bibliographystyle{IEEEtran}
\bibliography{refs}

\end{document}